\documentclass[letterpaper, 10 pt, conference]{ieeeconf}

\usepackage[lcgreekalpha,notext]{stix}
\usepackage{siunitx}

\usepackage{graphics} 
\usepackage{epsfig} 
\usepackage{graphicx}

\usepackage{amsmath}
\DeclareMathOperator*{\argmin}{arg\,min}
\usepackage{amssymb}
\usepackage[]{algorithm2e}

\newcommand{\matr}{\mathbfit}
\renewcommand{\vec}{\mathbfit}

\usepackage{url}
\usepackage{textcomp}
\usepackage{hyperref}

\usepackage{tabularx}
\usepackage{tabulary}
\usepackage{xfrac,nicefrac}
\usepackage{makecell}

\usepackage{multirow}
\newcolumntype{Y}{>{\centering\arraybackslash}X}
\usepackage{threeparttable}

\usepackage{colortbl}
\usepackage{xcolor}
\definecolor{sd}{gray}{0.5}
\definecolor{myblue}{rgb}{0.239215686,0.239215686,0.6}
\definecolor{mygreen}{rgb}{0.21568628,0.6901961,0.21568628}
\definecolor{myred}{rgb}{0.909803922,0.082352941,0.082352941}
\definecolor{mygray}{gray}{.9}
\definecolor{mypink}{rgb}{.99,.91,.95}
\usepackage{listings}

\usepackage{footnote}

\makeatletter
\let\NAT@parse\undefined
\makeatother
\usepackage[numbers]{natbib}

\usepackage{todonotes}
\usepackage{afterpage}

\providecommand{\U}[1]{\protect\rule{.1in}{.1in}}   
\IEEEoverridecommandlockouts
\begin{document}
\title{\LARGE \bf Inferring the Geometric Nullspace of Robot Skills\\ from Human Demonstrations}

\author{Caixia Cai$^1$, Ying Siu Liang$^{1,2}$, Nikhil Somani$^3$, Wu Yan$^1$
\thanks{Authors Affiliation: $^1$A*STAR Institute for Infocomm Research (I2R), Singapore, $^2$ A*STAR Human-centric AI (CHEEM) Programme, Institute of High Performance Computing, Singapore, $^3$A*STAR Advanced Remanufacturing and Technology Centre (ARTC), Singapore.}
\thanks{Email: ccxtum@gmail.com, liangys@ihpc.a-star.edu.sg, nikhil\_somani@artc.a-star.edu.sg, wuy@i2r.a-star.edu.sg}
}

\maketitle

\begin{abstract}
In this paper we present a framework to learn skills from human demonstrations in the form of geometric nullspaces, which can be executed using a robot. We collect data of human demonstrations, fit geometric nullspaces to them, and also infer their corresponding geometric constraint models. 
These geometric constraints provide a powerful mathematical model as well as an intuitive representation of the skill in terms of the involved objects.
To execute the skill using a robot, we combine this geometric skill description with the robot's kinematics and other environmental constraints, from which poses can be sampled for the robot's execution. 
The result of our framework is a system that takes the human demonstrations as input, learns the underlying skill model, and executes the learnt skill with different robots in different dynamic environments. We evaluate our approach on a simulated industrial robot, and execute the final task on the iCub humanoid robot.

\end{abstract}

\section{Introduction}
\label{sec:intro}

Imitating actions from demonstrations is an intuitive way to learn new tasks for both humans and robots.
Learning from Demonstration (LfD)~\cite{billard2008robot, Argall2009LFD} has been commonly used for human-robot skill-transfer, where the robot learns from examples provided by a human teacher. There are broadly two types of demonstrations: human performing the task, or human moving the robot through a joystick or kinesthetic methods to perform the task. 
The first approach has the advantage of being independent of any specific robot and may also be more intuitive or convenient for the human. The second way may be more suitable for tasks requiring higher precision (e.g. welding), where hardware independence is not important. In this work, we focus on the first approach and use a motion tracking system to track the positions of the human pose and objects in the scene. The learnt tasks are then executed on different types of robots.

Traditional LfD approaches either learn the exact trajectories or the low-level skill model by generalizing the demonstrated trajectories~\cite{calinon2007learning}. The modelling approach chosen to represent the skills greatly affects the variety of possible skills and their adaptability to different hardware and environments.
Constraint-based skill models offer a powerful and flexible choice, allowing us to model geometric constraints on the configuration and operational spaces \cite{Lin2017learning, perez2017c, somani2018constraint, subramani2018recognizing}, allowable velocities \cite{ortenzi2016kinematics}, and also forces and torques~\cite{Borghesan2016, decreICRA09, KressePhdThesis2017, subramani2018inferring,6942760}. Also, constraint-based approaches have proven to be amenable to semantic modelling using ontologies~\cite{Perzylo2015a} and can be used for intuitive programming interfaces for robotic tasks~\cite{Perzylo2016a}. In this work, we propose to use a constraint-based skill model with a LfD setup, where constraint-based descriptions and parameterized skill sequences are generated automatically from observing human demonstrations.

For example, when grasping a bottle, the hand pose is constrained to be at a certain distance and orientation with respect to the bottle.
These constraints define a space of valid poses for the robot to perform the task, which we refer to as the \textit{nullspace}.
To execute a task using a robot, we must additionally consider environment constraints and robot kinematics. 
For our example, the geometric nullspace of the task itself is a cylinder with a radius and height depending on the dimensions of the grasped object. To generate the grasp poses for a specific robot, we additionally consider the reachability of the robot's end-effector and possible obstacles in the robot's path.
The constraint-based description can be used to control the robot when executing a task and its parameters (e.g. height, radius of the bottle) can be edited to generalize to tasks with similar objects.

Learning task constraints is an active research area and has been addressed previously in relation with human demonstrations~\cite{vergara2019a, fang2016a, hayes2014}.
P\'erez-D'Arpino et al.~\cite{perez2017c} learn reaching and grasping tasks from demonstrations and describe them as a sequence of keyframes each associated with a set of geometric constraints.
Silverio et al.~\cite{silverio2018learning} use TP-GMM to simultaneously learn the operational and configuration space constraints (as well as priority hierarchies) from demonstrations.
Rodriguez et al.~\cite{rodriguez2008description} learn the relational positioning between two elements in space (point, line, plane) such as point-point distances, coincidences or parallelism.
Subramani et al.~\cite{subramani2018inferring} define six geometric constraints and infer them using both kinematic and force/torque information. 

However, current approaches to learn the nullspace and their geometric constraints are limited.
Most approaches use constraints that are handcrafted by an expert.
In our work, we propose an approach to learn geometric constraints and their nullspace only from kinematic information via human demonstrations.
This allows the robot to infer constraints for skills that have not been handcrafted by experts but are demonstrated by novice users.
In our experimental setup, we consider six basic skills that are demonstrated by the human. We use the data collected from these experiments to infer the geometric nullspace of the skill as well as their constraint-based description. These constraint-based descriptions can be adapted for different objects or environments by editing their parameters. To perform the skills on a robot, we combine the geometric constraints with constraints from the robot kinematic model and the environment and generate the required robot motions.
Finally, we execute the learnt skills on an iCub robot and show how the constraint nullspace can be used in a human-robot shared-control \cite{li2016} setting to maintain learnt constraints.

The key contribution of this paper is a novel LfD approach that infers a parameterized constraint-based skill model that is independent of the robot kinematics or environment. Through our experiments, we demonstrate how this approach can be used to easily learn new skills, adapt them to different objects/environments, execute them using different types of robots, and utilize the skill nullspace for human-robot shared control with physical interaction.

Section \ref{sec:framework} gives an overview of our framework and the basic skills.
In Section \ref{sec:constraint} we first describe the geometric constraint formulation and nullspace manifolds. Then we present the mathematical model for fitting the collected data to a nullspace manifold, and for mapping the nullspace to the geometric constraints.
Section \ref{sec:experiment} evaluates the proposed learning framework with a simulated industrial manipulator and on the iCub humanoid robot. The constraints and its nullspace of all the skills are  formulated. We present several real experiments (execution of learnt skills and a tea making task) on the iCub robot.
Finally, Section \ref{sec:conclusion} presents conclusions and future research directions.



\section{Learning from Demonstration Framework}
\label{sec:framework}

We propose a LfD framework to infer skills from human demonstrations that can be used for robot executions of similar tasks.
As LfD is an intuitive technique, it allows novice users to perform demonstrations of the skills.

\begin{figure}[tb]
    \centering
    \includegraphics[width=\linewidth]{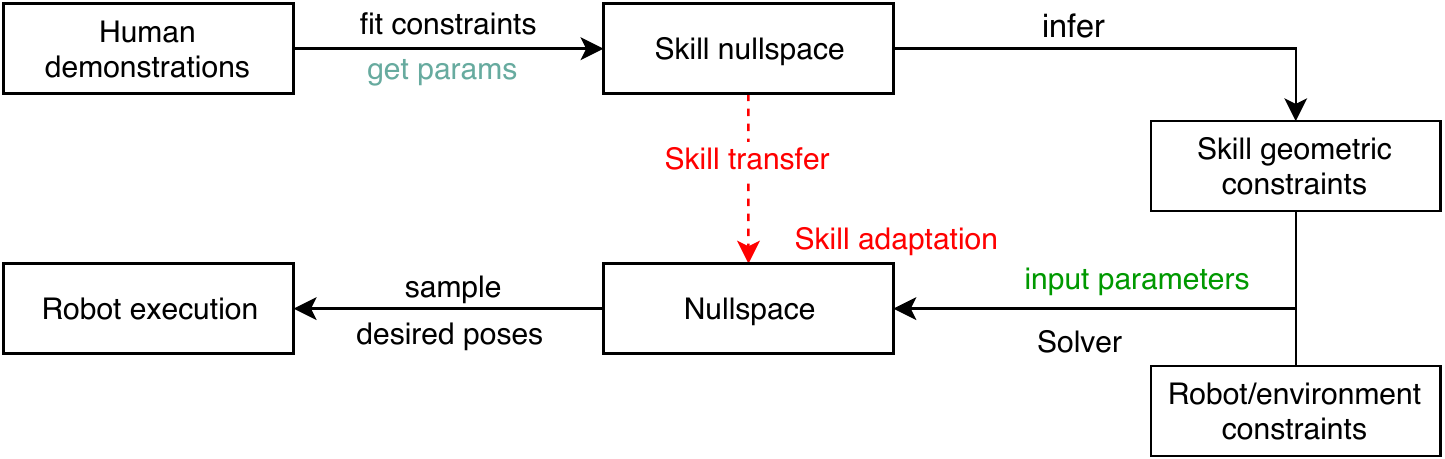}
    \caption{Overview of the proposed LfD framework: the demonstrated data of the skill is fit to the skill nullspace to infer geometric constraints, which is combined with robot/environment constraints for robot skill execution.}
    \label{fig_framework}
\end{figure}

\subsection{Framework}
Fig.~\ref{fig_framework} gives an overview of the proposed LfD framework and algorithm pipeline:
\begin{itemize}
    \item We first define a set of basic skills that we want to learn.
    \item We record samples of each skill from human demonstrations and collect kinematic information consisting of positions and orientations. 
    \item The set of data points in the skill demonstration represent samples in the nullspace of geometric constraints. We fit the data points to a nullspace manifold and estimate its parameters.
    \item Geometric constraints are inferred from the nullspace as a generalized representation of the learnt skills.
    \item The operator can easily modify the learnt skill by changing its parameters and adapt it to a different scenario: e.g. grasping a cup with different radius or height.
    \item The formulated geometric constraints are combined with robot/environment constraints.
    \item During robot execution, samples in the nullspace are generated by the constraint solver for executing the skill with a robot. In a dynamic environment, the poses are re-sampled in the nullspace to maintain the constraints.
\end{itemize}

\subsection{Data collection}
\begin{figure}[tb]
    \centering
    \includegraphics[width=\linewidth]{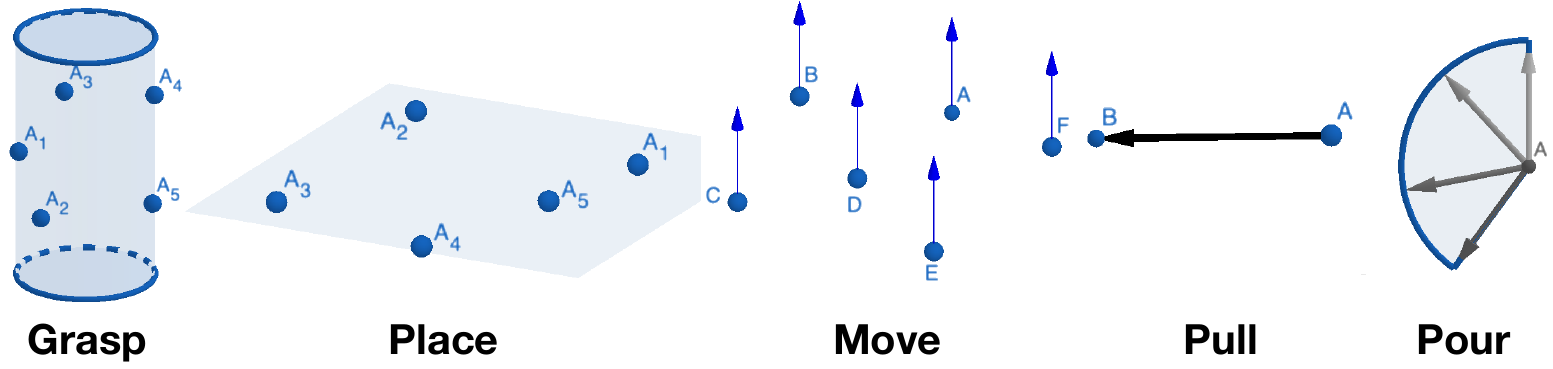}
    \caption{The collected demonstration data points for basic skills represent samples in the nullspace of their respective geometric constraints.}
    \label{fig_skill-overview}
\end{figure}

We define 6 basic skills and collect demonstration data (i.e., poses of all manipulation objects and 1 point on the wrist of the human hand) using OptiTrack 3D sensors\footnote{https://optitrack.com/}.
Some skills are discrete sample points (e.g. grasp) while others are continuous actions (e.g. pull). Hence, the data is collected either as individual poses (discrete) or trajectories (continuous): 

\begin{itemize}
    \item Grasp: grasping the bottle (discrete)
    \item Place: placing the bottle on the table (discrete)
    \item Move: moving the bottle/tray in 3D space (discrete)
    \item Pull: pulling a cup on the table closer (continuous)
    \item Pour: pouring the bottle into a cup (continuous)
    \item Mix: mixing a cup with a spoon (continuous)
\end{itemize}
The user performs multiple demonstrations of each skill individually (Fig.~\ref{fig_skill-overview}).

\section{Constraint-based skill model}
\label{sec:constraint}
Robot skills are implementations of behaviours or capabilities of the robot. In essence, they control motion of the robot in operational/configuration space in a way that this capability or behaviour is realized. In our formulation, they are expressed as constraints on the robot’s motion in the operational or configuration spaces.
In this constraint-based model, each skill is represented as a set of geometric inter-relational constraints between geometric entities in the setup, e.g. workpieces, tools. For example, grasping tasks can be represented using geometric constraints between the tools (or hand) and the object to be grasped. Therefore, the skills learnt through the geometric constraints and their nullspace are invariant to robots.

In this framework, the skill itself is described using \textit{\textbf{Geometric Constraints}}. Additional constraints from the environment such as collision avoidance and robot kinematics are added separately, and defined as \textit{\textbf{Environment Constraints}} and \textit{\textbf{Robot Constraints}} respectively. This separation of different constraint types allows hardware and environment agnostic skill descriptions that can be adapted to different robots and environments. The constraint modelling and solver used in this paper are based on our previous works~\cite{somani2018constraint, Somani2015b, Somani2016a, Somani2017a}.

\begin{table}[tb]
	\centering
	\caption{Geometric constraint models}
	\begin{tabular*}{\linewidth}{l @{\extracolsep\fill} l @{\extracolsep\fill} l @{\extracolsep\fill} l}
		\noalign{\smallskip}
		\hline
		\noalign{\smallskip}
		\textbf{Fixed} & \textbf{Constr.} & \text{\thead{Geometric \\ Constraint}} & \text{\thead{Transformation Manifold \\ (Translation, Rotation)}}\\
		\noalign{\smallskip}
		\hline
		\noalign{\smallskip}
		Line$_{1}$ & Point$_{2}$ & Distance & Cylinder, $\textbf{SO(3)}$ \\
		Line$_{1}$ & Line$_{2}$ & Distance & Cylinder, OneParallel \\
		\noalign{\smallskip}
		Line$_{1}$ & Line$_{2}$ & Angle & $\mathbb{R}^3$, OneAngle\\
		\noalign{\smallskip}
		Plane$_{1}$ & Point$_{2}$ & Distance & Plane, $\textbf{SO(3)}$\\
		\noalign{\smallskip}
		Plane$_{1}$ & Line$_{2}$ & Distance & Plane, OneAngle \\
		\noalign{\smallskip}
		Plane$_{1}$ & Line$_{2}$ & Angle & $\mathbb{R}^3$, OneAngle\\
		\noalign{\smallskip}
		Plane$_{1}$ & Plane$_{2}$ & Distance & Plane, OneParallel\\
		\noalign{\smallskip}
		Plane$_{1}$ & Plane$_{2}$ & Angle & $\mathbb{R}^3$, OneAngle\\
		\noalign{\smallskip}
		\noalign{\smallskip}
		\hline
		\noalign{\smallskip}
	\end{tabular*}
\label{table_geometric_constraint_decomposition}
\end{table}
\subsection{Geometric constraints}

A geometric constraint is defined as a geometric relation between two shapes (fixed and constrained) that affects the relative transformation of the constrained shape w.r.t. the fixed shape. These can relate to distances, angles, parallelism or coincidences between  points, lines or planes. The set of relative transformations that satisfy the constraints is the nullspace of the geometric constraint. Table~\ref{table_geometric_constraint_decomposition} shows some of the geometric constraint models and their corresponding nullspaces that we used in this work.
\subsubsection*{\textbf{Constraint Nullspace}}
As seen from Table~\ref{table_geometric_constraint_decomposition}, the nullspace of geometric constraints contains two parts describing a transformation manifold: rotation and translation. Tables~\ref{table_shape_translation_submanifold_model} and \ref{table_shape_rotation_submanifold_model} show the parametric models and projection functions for the translation and rotation manifolds respectively. 

\subsection{Fitting Data to a Skill Nullspace Manifold}
The experimental data collected from our teaching experiments contains poses of the different entities in the scene, e.g. bottle, cup, hand. For each entity, we manually define a set of geometric shapes that approximate its geometry. This decomposition can also be done automatically using CAD models~\cite{Perzylo2015a} or primitive fitting~\cite{Somani2015c}. Each of these geometries is then tracked as a frame in the data recordings. The recordings are used to calculate relative transformations between the different frames. The set of data points ($[\vec{u}, \matr{R}] \in \mathbb{D}, \mathbb{D} \in \mathbf{SE}(3)$) in the skill demonstration recording represent samples in the nullspace of geometric constraints. In case of continuous actions, the temporal spread of these data points is also important as they represent a trajectory in the nullspace of geometric constraints.
We use these samples to classify the nullspace type and fit the data points to estimate its parameters. This is done by formulating this fitting problem as a non-linear optimization problem (Eq.~\ref{eq:optimization-problem}), where the parameters of the nullspace model $\vec{\phi}=[\vec{\phi}_T, \vec{\phi}_R]$ are the optimization variables and the optimization function is the distance of the data points from the fitted nullspace model~(Eq.~\ref{eq:distanceT}, Eq.~\ref{eq:distanceR}).

It is possible to have multiple nullspace definitions that fit the data. To avoid over-generalization, we consider the complexity of different models in the fitting algorithm and chose the most restrictive model that can fit the data. For example, a set of points lying on a line will also fit correctly on a plane. We evaluate the fitting error for simpler models and progressively move to more complex models until a threshold value of the fitting error is reached. In our examples, the order followed was Point, Line, Plane, Cylinder.

\begin{equation}
	\label{eq:optimization-problem}
	\begin{aligned}
		\vec{\phi}_{opt} = \argmin\limits_{[\vec{\phi}_T, \vec{\phi}_R]} 
		\sum_{[\vec{u},\matr{R}]_i \in \mathbb{D}}{(\operatorname{distT}(\vec{\phi}_T, \vec{u}) + \operatorname{distR}(\vec{\phi}_R,\vec{v}_c,\matr{R}))}\\
	\end{aligned}
\end{equation}
\begin{equation}
	\label{eq:distanceT}
	\begin{aligned}
		\operatorname{distT}(\vec{\phi}_T, \vec{u}) = \lVert \operatorname{ProjT}(\vec{\phi}_T, \vec{u})-\vec{u}\rVert _{2} \\
	\end{aligned}
\end{equation}
\begin{equation}
	\label{eq:distanceR}
	\begin{aligned}
		\operatorname{distR}(\vec{\phi}_R, \vec{v}_c, \matr{R}) = \lVert \operatorname{quat}(\operatorname{ProjR}(\vec{\phi}_R,\vec{v}_c, \matr{R}))-\operatorname{quat}(\matr{R})\rVert _{2}\\
	\end{aligned}
\end{equation}

\begin{table}[tb]
	\centering
	\begin{threeparttable}
	\caption{Parametric representation and projection operations for translation manifolds}
 	\begin{tabular*}{\linewidth}{l @{\extracolsep\fill} l @{\extracolsep\fill} l}
		\noalign{\smallskip}
		\hline
		\noalign{\smallskip}
		\textbf{Manifold} & \textbf{Parameters $\vec{\phi}_T$}\hphantom{--} & \textbf{Projection of Point $\operatorname{ProjT}(\vec{\phi}, \vec{u})$}\\
		\hline
		\noalign{\smallskip}
		\noalign{\smallskip}
		$\mathbb{R}^3$& - & $\vec{u}$\\ 
		\noalign{\smallskip}
		Point & [$\vec{p}$] & $\vec{p}$\\
		\noalign{\smallskip}
		Line & [$\vec{p},\hat{\vec{a}}$] & $\vec{p}+((\vec{u}-\vec{p})\cdot\hat{\vec{a}})\hat{\vec{a}}$\\
		\noalign{\smallskip}
		Circle & [$\vec{p},\hat{\vec{n}},r$] & $\vec{p}+r[\operatorname{ProjT}_{\mathrm{Plane}(\vec{p},\hat{\vec{n}})}(\vec{u})-\vec{p}]\tnote{1}$\\
		\noalign{\smallskip}
		Plane & [$\vec{p},\hat{\vec{n}}$] & $\vec{u}+((\vec{p}-\vec{u})\cdot\hat{\vec{n}})\hat{\vec{n}}$\\
		\noalign{\smallskip}
		Cylinder & [$\vec{p},\hat{\vec{a}},r,h$] & $\operatorname{ProjT}_{\mathrm{Plane}(\vec{p},\hat{\vec{a}})}(\vec{u})+r[\vec{u}-\operatorname{ProjT}_{\mathrm{Plane}(\vec{p},\hat{\vec{a}})}(\vec{u})]\tnote{1}$\\
		\noalign{\smallskip}
		\hline
		\noalign{\smallskip}
	\end{tabular*}
	\label{table_shape_translation_submanifold_model}
	\begin{tablenotes}
	\item[1] vectors enclosed in $[]$ are considered normalized
	\end{tablenotes}

	\end{threeparttable}
\end{table}

\begin{table}[t]
	\centering
	\begin{threeparttable}
	\caption{Parametric representation and projection for rotation manifolds}
	\begin{tabular*}{\linewidth}{l @{\extracolsep\fill} l @{\extracolsep\fill} l}
		\noalign{\smallskip}
		\hline
		\noalign{\smallskip}
		\textbf{Manifold} & \textbf{Parameters $\vec{\phi}_R$} & \textbf{Projection of Vector $\operatorname{ProjR}(\vec{\phi}, \vec{v}_{c}, \matr{R}_i)$}\\
		\noalign{\smallskip}
		\hline
		\noalign{\smallskip}
		$\mathbf{SO(3)}$ & - & $\matr{R}_i$\\
		\noalign{\smallskip}
		OneParallel & [$\vec{v}_{f}$] & $\matr{R}_{AA}(\vec{w}_R, \alpha_R)~\matr{R}_i$\\
		\noalign{\smallskip}
		OneAngle & [$\vec{v}_{f}, \theta$] & $\matr{R}_{AA}(\vec{w}_R, \alpha_R-\theta)~\matr{R}_i$\\
		\noalign{\smallskip}
		\hline
		\noalign{\smallskip}
	\end{tabular*}
	\label{table_shape_rotation_submanifold_model}
	\begin{tablenotes}
	\item[1] $\vec{v}_f$ = Fixed vector, $\vec{v}_c$ = Constrained vector, $\vec{w} = \vec{v}_{f}\times\vec{v}_{c}$, $\alpha = \angle(\vec{v}_{f},\vec{v}_{c})$, $\alpha_R = \angle(\vec{v}_{f},\vec{v}_{c})$, $\vec{w}_R = \vec{v}_{f}\times\vec{v}_{c}$, $\matr{R}_{AA}(\vec{a}, \theta)$ = Rotation matrix defined by axis $\vec{a}$ and angle $\theta$, $\matr{R}_i$ = Input rotation matrix
	\end{tablenotes}
	\end{threeparttable}
\end{table}

\subsection{Mapping the Skill Nullspace to Geometric Constraints}
Using the non-linear optimization approach, the skill nullspace can be estimated. 
From this nullspace definition, samples can be generated for executing the skill with a robot. However, this nullspace description does not provide much semantic information about the skill itself. It can be difficult for a person to understand the learnt skill from the nullspace description alone. A skill description in terms of geometric relations (e.g. coincident, concentric) between geometric entities present in the scene (e.g. plane of the table, cylinder of the cup) are much more intuitive for the operator to understand. 
The operator can also easily modify the learnt skill by changing its parameters and adapt it to a different scenario, e.g. grasping a cup with a different radius or height.
To map the geometric nullspace with the corresponding constraints, we use the mapping in Table~\ref{table_geometric_constraint_decomposition} along with the list of geometries that form the different objects (e.g. Plane$_{\text{table}}$, Cylinder$_{\text{cup}}$).

\subsection{Executing learnt skills using a robot}
Once the constraint-based definitions of the learnt skill are obtained, they can be solved to generate the required robot motions for execution.

\subsubsection{\textbf{Robot and Environment Constraints}}
In addition to direct geometric relations that define the skill, the robot controller needs to consider the kinematic structure of the robot (e.g. 6-DoF industrial robot, 53-DoF iCub), incorporate safety requirements arising from the robot model (e.g. joint limits) and constraints from the environment (e.g. collision avoidance). Furthermore, there are additional limitations on the robot's speed and acceleration when a human is present in the robot's workspace to ensure safe collaboration. All of these aspects are also modeled as non-linear constraints, based on our previous work~\cite{Somani2016a}.

\subsubsection{\textbf{Constraint Solver}}
We use a hybrid constraint-solving approach~\cite{somani2018constraint}, which is a combination of the exact solver for geometric constraints~\cite{Somani2017a} followed by an iterative solver~\cite{Somani2016a} that combines the robot and environment constraints to generate a non-linear optimization problem. This optimization problem is solved using the NLOpt library~\cite{nlopt} to generate the Cartesian or joint positions required to execute the task on a specific robot.

\subsubsection{\textbf{Task Priorities}}
When there are multiple constraints from different aspects that the robot must consider, the definition of priorities is important to ensure safe execution of learnt skills. In our case, we consider multiple priority levels. The highest priority is assigned to the robot joint limitations and collision avoidance. The second priority is assigned to the geometric constraints describing the task. The lowest priority levels are for maintaining an optimal (or most reachable) posture and minimizing the distance between successive joint positions.

\section{Experiment Evaluation}
\label{sec:experiment}
In this section, we show how our LfD framework is used in robotics applications. We demonstrate several typical robotic skills learnt using our framework and represented as geometric constraints, e.g. grasp, move and place objects. The formulated geometric constraints together with robot and environment constraints can be solved by constraint solvers to provide the desired poses for the robot execution. 

As mentioned earlier, the skills learnt from human demonstration are invariant to the robot. 
Therefore, the proposed framework is evaluated on different robots. Firstly, we tested the execution of learnt skills on a 6-DoF industrial robot in simulation. Furthermore, we also evaluated it on the 53-DoF iCub humanoid robot, learning six basic skills and a tea making task in a human-robot interaction scenario.

\subsection{Robot Skills}

We present several examples of robot skills with their constraint-based definition as well as an analysis of the nullspace of each skill. We test all the skills and illustrate their spaces on a 6-DoF industrial robot.\footnote{See \textit{demo 1} of the video: \href{https://youtu.be/8DFZG8qrwYA}{https://youtu.be/8DFZG8qrwYA} }

\begin{figure}[thpb]
        \centering
	\includegraphics[width=.68\linewidth]{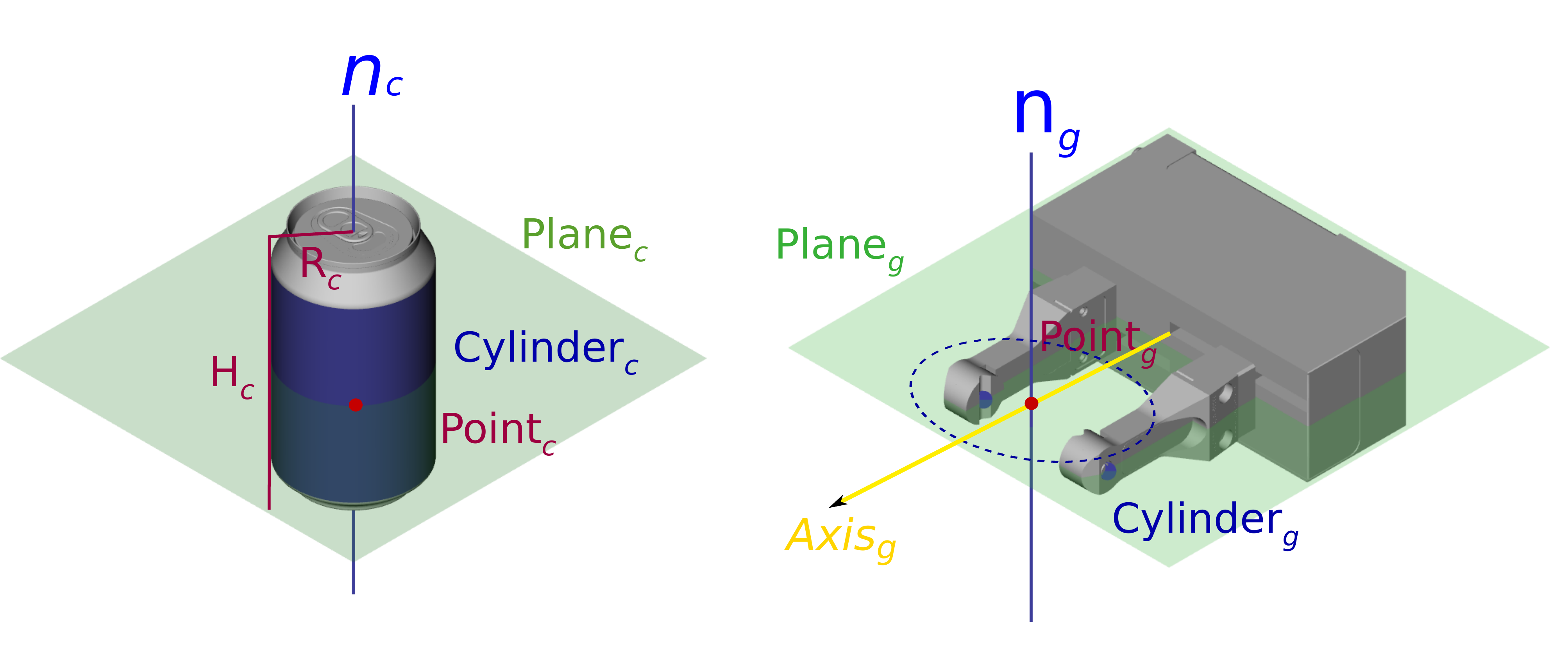}
	\includegraphics[width=0.3\linewidth]{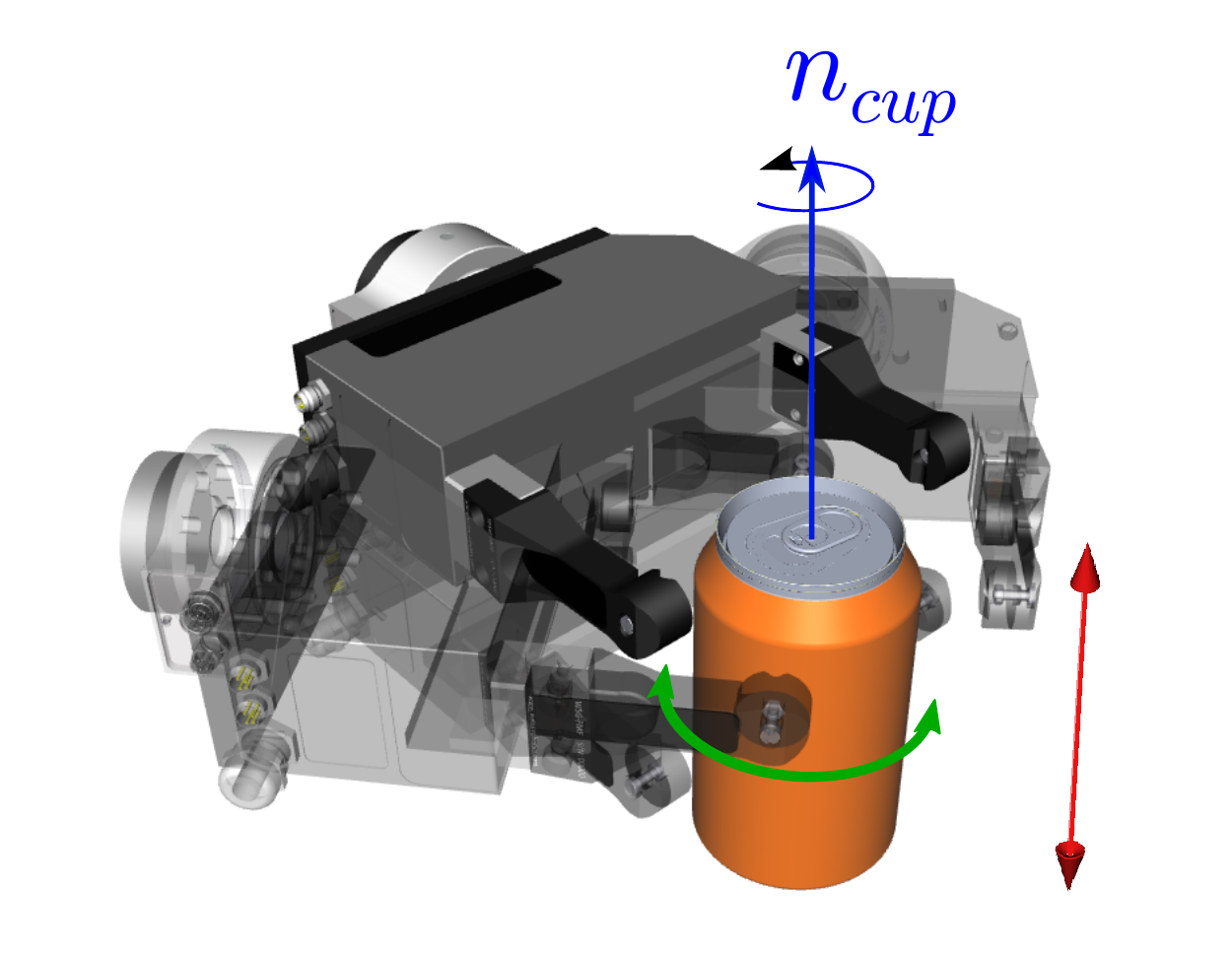}	
	\smallskip
	\footnotesize
	\begin{tabularx}{0.65\linewidth}{Y}
		\noalign{\smallskip}\hline\noalign{\smallskip}
		\textbf{Grasping - Geometric Constraints}\\
		\noalign{\smallskip}\hline\noalign{\smallskip}
		$ \operatorname{Concentric}({\color{myblue}\text{Cylinder}_c} , \ {\color{myblue}\text{Cylinder}_{g}}) $\\
		\noalign{\smallskip}
		$ \nicefrac{-{\color{myred}\text{H}_{c}}}{2} \leq \operatorname{Distance}({\color{mygreen}\text{Plane}_c} , \ {\color{mygreen}\text{Plane}_{g}}) \leq \nicefrac{{\color{myred}\text{H}_{c}}}{2}$\\
		\noalign{\smallskip}
	
		\noalign{\smallskip}\hline\noalign{\smallskip}		
		\textbf{Skill Nullspace}\\
		\noalign{\smallskip}\hline\noalign{\smallskip}		
		A cylinder + Rotation around  ${\color{myblue}\text{n}_{c}}$ 
		\\
		\noalign{\smallskip}\hline\noalign{\smallskip}		
	\end{tabularx}	
	\caption{A cup grasping task expressed using geometric constraints with inequalities ($H_c$, $R_c$ and $n_c$ are the height, radius and axis of the cup), resulting in the skill nullspace (right).}
	\label{fig_grasp}
\end{figure}


\subsubsection{\textbf{Grasping a cup}}
This skill involves grasping a cylindrical object (a cup). The constraint-based formulation of this skill is shown in Fig.~\ref{fig_grasp} with some sample target poses for grasping and the nullspace of its geometric constraints. 
The relative position of the cup and gripper in the grasp skill are underspecified, with one DoF as the rotation along the cylinder's axis of cup ${\color{myblue}\text{n}_c}$.
The translation manifold of this skill nullspace is a cylinder. The grasp poses lie on the cylinder with freedom along the translation axis ${\color{myblue}\text{n}_c}$ within the height $\color{myred}\text{H}_{c}$ and rotation around axis ${\color{myblue}\text{n}_c}$.


\subsubsection{\textbf{Other skills}}
\begin{figure*}[tb]
	\centering	
 	\includegraphics[width=0.7\linewidth]{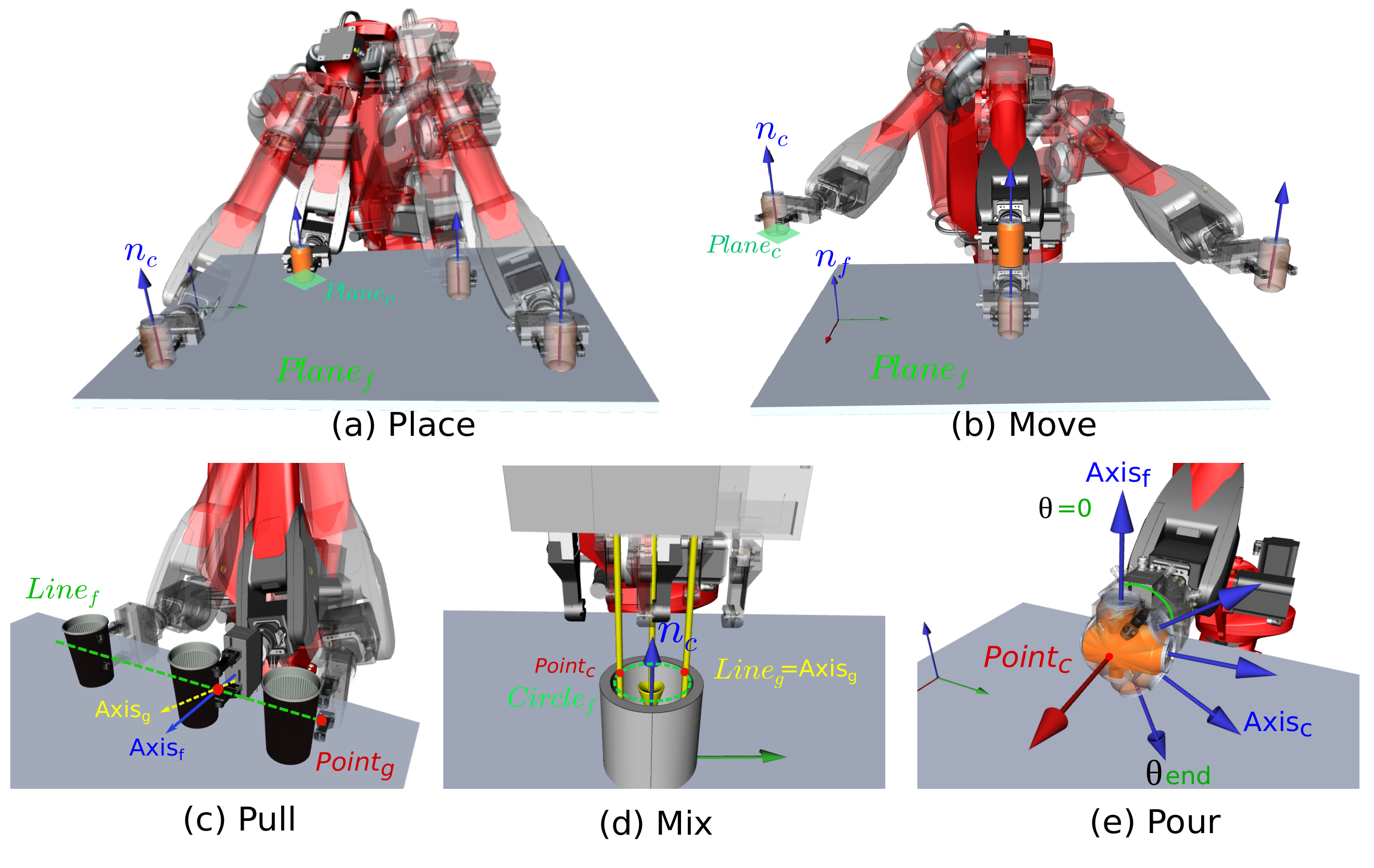}%
 	\\
 	\smallskip
	\begin{tabular*}{\linewidth}{l @{\extracolsep\fill} c @{\extracolsep\fill} c}
		\noalign{\smallskip}
		\hline
		\noalign{\smallskip}
		\textbf{Skill} & \textbf{Geometric Constraints} & \textbf{Skill Nullspace}\\
		\noalign{\smallskip}
		\hline
		\textbf{Place} & $ \operatorname{Coincident}({\color{mygreen}\text{Plane}_c} , \ {\color{mygreen}\text{Plane}_f})$ & \text{\thead{Rotation: ${\operatorname{OneParallelManifold}(\color{myblue}\text{n}_f)}$\\
		Translation: $\color{mygreen}\text{Plane}_f$}}\\
		\textbf{Move} & $ \operatorname{Parallel}({\color{mygreen}\text{Plane}_c} , \ {\color{mygreen}\text{Plane}_f})$ & \text{\thead{Rotation: ${\operatorname{OneParallelManifold}(\color{myblue}\text{n}_f)}$\\
		Translation: $\mathbb{R}^3$}}\\
		\textbf{Pull} & \text{\thead{$\operatorname{Coincident}({\color{myred}\text{Point}_g} , \ {\color{myblue}\text{Point}_f}),\ \ {\color{myblue}\text{Point}_f} \in {\color{mygreen}\text{Line}_f}$ \\ $ \theta_{min} \leq \operatorname{Angle}( {\color{myblue}\text{Axis}_g},  \ {\color{myblue}\text{Axis}_f}) \leq \theta_{max} $\\
		(Angle along ${\color{myblue}\text{n}_g}$)}} & \text{\thead{Rotation: $\operatorname{OneAngleManifold}({\color{myblue}\text{Line}_g}$, $\theta_{min}$, $\theta_{max}$)\\
		Translation: $\color{mygreen}\text{Line}_f$}}\\
		\textbf{Mix} & \text{\thead{$\operatorname{Coincident}({\color{myred}\text{Point}_c} , \ {\color{myblue}\text{Point}_f}),\ \ {\color{myblue}\text{Point}_f} \in {\color{mygreen}\text{Circle}_f} $\\
		$ \operatorname{Distance}({\color{myblue}\text{Line}_g} , \ {\color{myblue}\text{n}_c})$}} & \text{\thead{Rotation: ${\operatorname{OneParallelManifold}(\color{myblue}\text{n}_c)}$\\
		Translation: ${\color{mygreen}\text{Circle}_f}$
		}}\\
		\textbf{Pour} & \text{\thead{$ \operatorname{Coincident}({\color{myred}\text{Point}_c} , \ {\color{myblue}\text{Point}_f}) $\\
		$ \operatorname{Angle}({\color{myblue}\text{Axis}_c} , \ {\color{myblue}\text{Axis}_f}) = \theta, 0 \leq \theta \leq \theta_{end} $
		}} & \text{\thead{Rotation: $\operatorname{OneAngleManifold}({\color{myblue}\text{Axis}_c}$, $0$, $\theta_{end}$)\\
		Translation: --}}\\
		\hline
	\end{tabular*}
	\caption{Geometric constraints and nullspaces for the learnt skills described with their transformation manifolds and inequalities where necessary.}
	\label{fig_other_constraints}
\end{figure*}

Besides grasp skills, we also learnt several other skills. Fig.~\ref{fig_other_constraints} lists the constraint-based formulation of the skills and their nullspace of geometric constraints. 

\begin{itemize}
 \item \textbf{Place:} Place an object on a plane. The geometric constraint for this skill is that ${\color{mygreen}\text{Plane}_c}$, the bottom plane of the cup is coincident with ${\color{mygreen}\text{Plane}_f}$, the table plane. The nullspace of this skill is a two DoF translation along ${\color{mygreen}\text{Plane}_f}$, and one DoF rotation along the normal direction of ${\color{mygreen}\text{Plane}_f}$. Fig.~\ref{fig_other_constraints}a illustrates some sample poses in this nullspace. 
 
 \item \textbf{Move:} This skill involves moving a cup or a tray containing liquid, where the axis ${\color{myblue}\text{n}_c}$ should be parallel to ${\color{myblue}\text{n}_f}$ to avoid spilling 
(see Fig.~\ref{fig_other_constraints}b).
 
 \item \textbf{Pull:}  For \textit{pull}, the center point ${\color{myred}\text{Point}_g}$ of the gripper must move along the desired line ${\color{mygreen}\text{Line}_f}$ between a start and end point. During this movement, the orientation of the gripper is free around one axis and defines the nullspace of the skill.  Fig.~\ref{fig_other_constraints}c shows some sample target poses for the gripper and also illustrates the nullspace of its geometric constraints.
  
 \item \textbf{Mix:}  The desired trajectory of this task is a ${\color{mygreen}\text{Circle}_f} $. The tip of the mixing tool follows this desired trajectory while the tool line, which is parallel to the gripper $\text{Axis}_g$, maintains a fixed distance (radius) to the axis of the cup ${\color{myblue}\text{n}_c}$. Fig.~\ref{fig_other_constraints}d shows some sample poses.

 \item \textbf{Pour:} This skill is a pure rotation movement around one axis based on one point on the cup. The rotation is between a starting angle and an ending angle. Some sample target rotations and also the nullspace of its geometric constraints are illustrated in Fig.~\ref{fig_other_constraints}e.
\end{itemize}

Similarly, more basic robot skills and tasks can be formulated using geometric constraints and learnt using our framework, e.g. writing, erasing and so on.

\begin{figure*}[tb]
	\centering	
 	\includegraphics[width=0.75\linewidth]{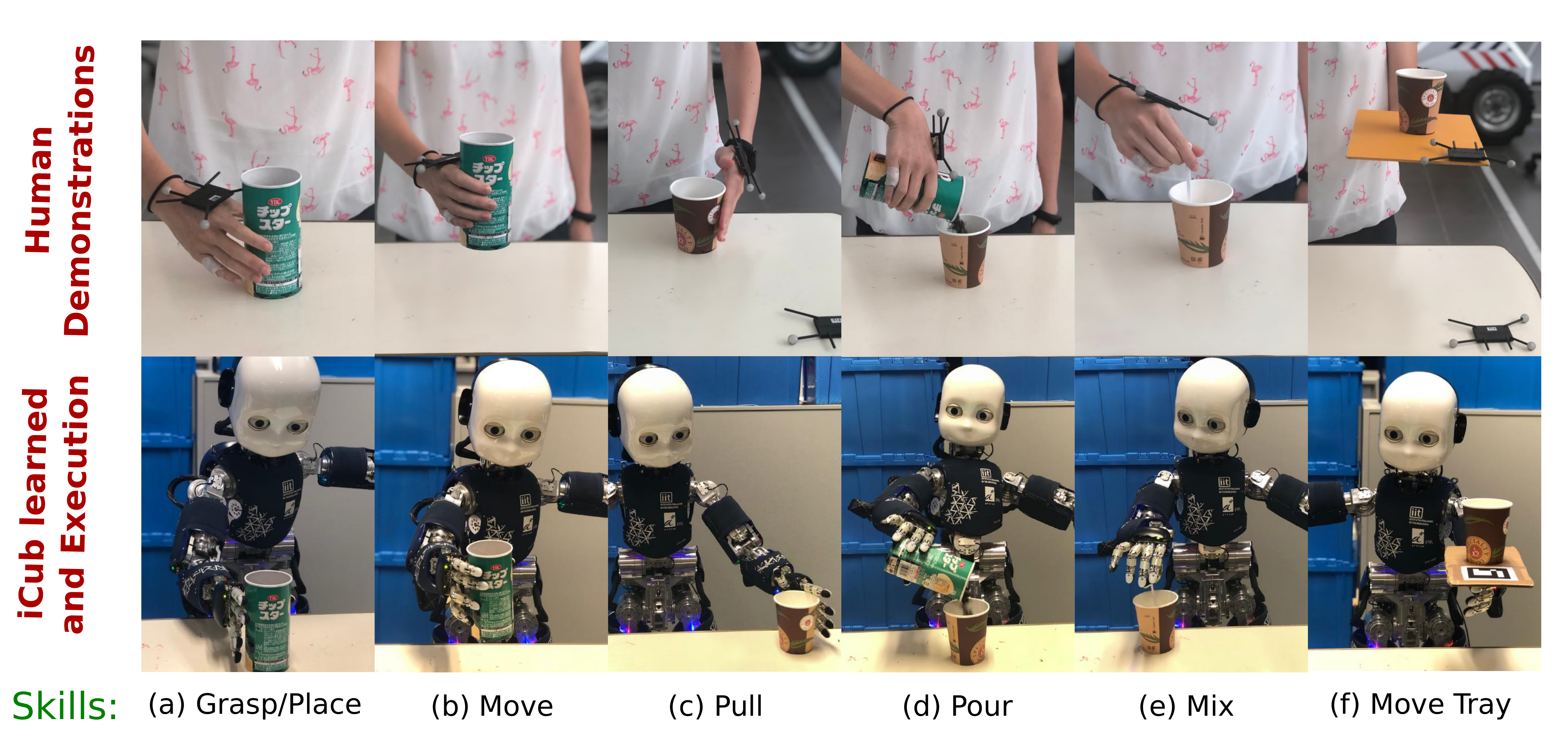}%
 	\caption{iCub learnt the skills from human demonstrations based on geometric constraints and skill nullspace.}
	\label{fig_icub_skills}
\end{figure*}

\subsection{Skill Execution on iCub}
Our experimental platform is the open-source iCub humanoid robot with 53 DoF. We conducted experiments on iCub with 7 DoF on each arm and 9 DoF on each hand. In our framework, the robot learns the skills as geometric constraints. Solutions in the nullspace of these constraints are desired poses for the robot to perform this learnt skill.

Fig.~\ref{fig_icub_skills} shows six basic robot skills in a human-robot interaction scenario that are learnt by iCub using our framework.
The pictures in the first row show the human demonstrations of each skill. After learning the geometric constraints and nullspaces, iCub is able to perform these skills adaptively, with different parameters (e.g. different objects position, different object shape parameters).

\subsection{Skills Sequence}
In this section, we demonstrate a tea/coffee making task, performed by the iCub using the learnt skills
. The task contains the following sequence of skills:

\begin{enumerate}
 \item \textbf{Pull}: \  As the cup is too far, iCub first needs to pull the cup nearer to the center of the table.
 \item \textbf{Grasp + Pour + Place}: \ iCub grasps the tea bottle, pours tea leaves into the cup, and then places the bottle back on the table.
 \item \textbf{Grasp + Pour + Place}: \ iCub grasps the water bottle, pours water into the cup, then places the bottle back.
 \item \textbf{Mix}:  \ After putting the tea/coffee and water into the cup, iCub mixes them with a spoon.
 \item \textbf{Move (Move Tray)}: \ The tea/coffee is prepared and iCub will grasp it and offer to the human directly or put it on a tray first then offer to human.  In this step, the move skill should keep the cup upright since there is liquid inside the cup.
\end{enumerate}

\subsection{Prioritized Task Control}


We perform other tasks together with learnt skills using our optimization-based prioritized task control. 



\begin{figure}[thpb]
        \centering
	\includegraphics[width=.95\linewidth]{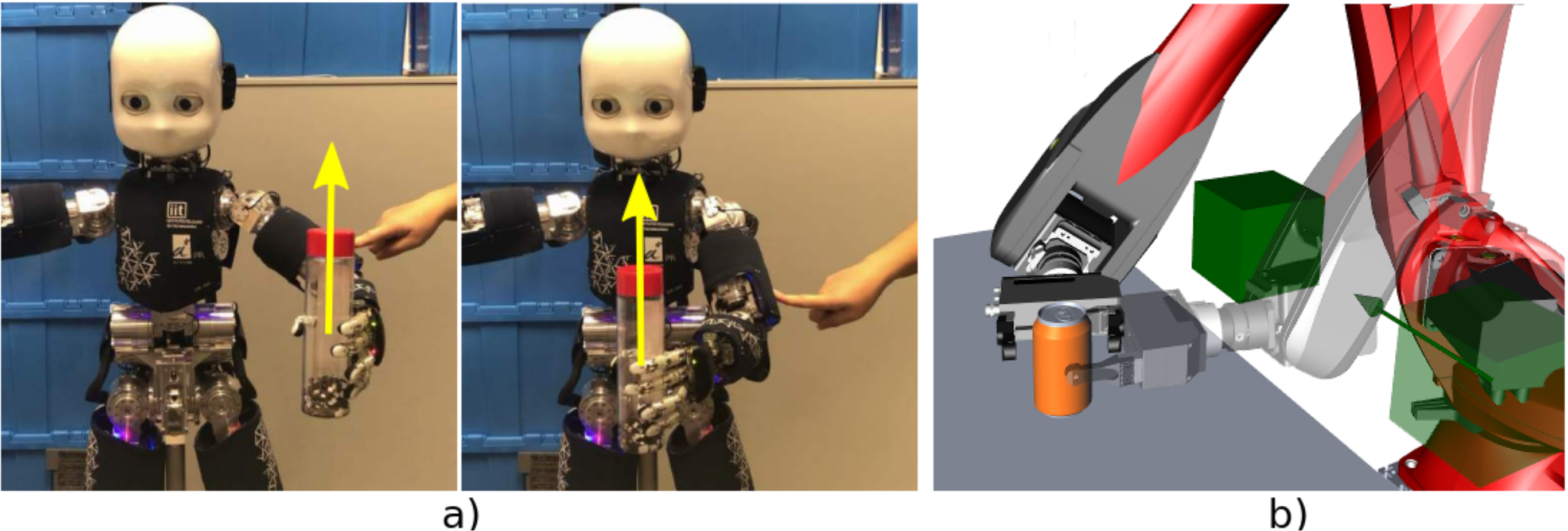}
	\caption{(a) iCub keeps constraint (cup upright) when interacting with human
	(b) cup grasping with obstacle avoidance using a 6-DoF robot.}
	\label{fig_interaction}
\end{figure}

\begin{enumerate}
\item \textbf{Cup grasping with obstacle avoidance:\\}
The obstacle avoidance task is at a higher priority, and our constraint solver~\cite{Somani2016a} tries to find an optimal solution that avoids the collision, while also adjusting the grasp position within the learnt nullspace of the \textbf{Grasp} skill (see Fig.~\ref{fig_interaction}b)

\item \textbf{Compliant interaction with cup upright:\\}
In this task, the human is interacting with iCub using a compliant mode, combining force and motion constraints into a prioritized task controller~\cite{Cai2015ROBIO}.\footnote{See also \textit{demo 2 - part 3} of the video: \href{https://youtu.be/8DFZG8qrwYA}{https://youtu.be/8DFZG8qrwYA} } 
The constraint is that the cup held by the iCub has to be kept upright during the interaction since there is liquid inside the cup. We define the higher priority task as the learnt \textbf{Move} skill, which is keeping the orientation of the cup upright. Then the compliant interaction is performed in the nullspace of the \textbf{Move} skill. Fig.~\ref{fig_interaction}a shows two different moments during the interaction, where we can see the cup is always kept upright.
\end{enumerate}

\section{Conclusions}
\label{sec:conclusion}
This paper presents a Learning from Demonstration (LfD) framework based on a constraint-based skill model, where a set of parameterized geometric constraints representing the skill can be inferred from human demonstrations. We use human demonstrations of basic skills to collect data about the kinematics information between different objects involved in the skill. We developed an algorithm to fit this data to a geometric nullspace manifold and also infer the skill's underlying geometric constraints. The inferred geometric constraints provide semantically rich information about the skill and can be used to adapt it to different kinds of robots and objects. These geometric task constraints are combined with the robot's and environment constraints, and solved together to generate target motions for the robot.
The proposed framework is evaluated on both a simulated industrial manipulator and the iCub humanoid robot.

In our experiments, we have considered six basic skills. For this purpose, the simple non-linear least squares regression to fit the data was sufficient. This design was based on our observation that constraint-based approaches with a limited set of primitive shapes can be powerful in describing a large variety of engineering models and corresponding robotic tasks. However, with more complex skills and constraint models, a more powerful algorithm for data fitting may be required. Moreover, bringing the deep learning methods to cover the whole LfD pipeline from human observation to inference of the geometric constraints would also be interesting for future work.

\section*{Acknowledgements}
This research is partially supported by the Agency for Science, Technology and Research (A*STAR) under its AME Programmatic Funding Scheme (Project \#A18A2b0046).



\bibliography{main.bib}
\bibliographystyle{IEEEtran}

\end{document}